\documentclass[letterpaper]{article} % DO NOT CHANGE THIS
\usepackage{aaai25}  % DO NOT CHANGE THIS
\usepackage{times}  % DO NOT CHANGE THIS
\usepackage{helvet}  % DO NOT CHANGE THIS
\usepackage{courier}  % DO NOT CHANGE THIS
\usepackage[hyphens]{url}  % DO NOT CHANGE THIS
\usepackage{graphicx} % DO NOT CHANGE THIS
\usepackage{natbib}  % DO NOT CHANGE THIS AND DO NOT ADD ANY OPTIONS TO IT
\usepackage{caption} % DO NOT CHANGE THIS AND DO NOT ADD ANY OPTIONS TO IT
\usepackage{algorithm}
\usepackage{algorithmic}
\usepackage{newfloat}
\usepackage{listings}
\DeclareCaptionStyle{ruled}{labelfont=normalfont,labelsep=colon,strut=off} % DO NOT CHANGE THIS
\floatstyle{ruled}
\newfloat{listing}{tb}{lst}{}
\floatname{listing}{Listing}
\usepackage{multirow}
\usepackage{amssymb}
\usepackage{amsmath}
\usepackage{booktabs}

\title{Capture Global Feature Statistics for One-Shot Federated Learning}
\author{
    Zenghao Guan\textsuperscript{\rm 1, 2, 3},
    Yucan Zhou\textsuperscript{\rm 1, 3}\footnote{Corresponding author},
    Xiaoyan Gu \textsuperscript{\rm 1, 2, 3}\footnotemark[1]
}
\affiliations{
\textsuperscript{\rm 1}Institute of Information Engineering, Chinese Academy of Sciences\\
\textsuperscript{\rm 2}School of Cyber Security, University of Chinese Academy of Sciences\\
\textsuperscript{\rm 3}State Key Laboratory of Cyberspace Security Defense \\
zenghaoguan.cs@gmail.com, \{zhouyucan, guxiaoyan\}@iie.ac.cn
}

\usepackage{bibentry}
\begin{document}

\maketitle

\begin{abstract}
Traditional Federated Learning (FL) necessitates numerous rounds of communication between the server and clients, posing significant challenges including high communication costs, connection drop risks and susceptibility to privacy attacks. One-shot FL has become a compelling learning paradigm to overcome above drawbacks by enabling the training of a global server model via a single communication round. However, existing one-shot FL methods suffer from expensive computation cost on the server or clients and cannot deal with non-IID (Independent and Identically Distributed) data stably and effectively. To address these challenges, this paper proposes \textbf{FedCGS}, a novel \textbf{Fed}erated learning algorithm that \textbf{C}apture \textbf{G}lobal feature  \textbf{S}tatistics leveraging pre-trained models. With global feature statistics, we achieve training-free and heterogeneity-resistant one-shot FL. Furthermore, we extend its application to personalization scenario, where clients only need execute one extra communication round with server to download global statistics. Extensive experimental results demonstrate the effectiveness of our methods across diverse data heterogeneity settings. Code is available at \url{https://github.com/Yuqin-G/FedCGS}.
\end{abstract}


\section{Introduction}
\label{sec:intro}

Federated Learning (FL) is an emerging framework that enables multiple parties to participate in collaborative learning under the coordination of a central server, which aggregates model updates rather than private data, enhancing privacy in distributed learning \cite{McMahan17}. However, typical FL requires numerous communication rounds between the server and clients, leading to significant challenges. First, clients must maintain constant connections with the server to upload and receive updates, resulting in high communication costs and the risk of connection drops \cite{li2020federated2, kairouz2021advances, chen2023workie}, which is unbearable for bandwidth-limited or real-time FL applications. Second, frequent communication increases the system's vulnerability to data and model poisoning attacks \cite{mothukuri2021survey, rao2024privacy, guan2024gie, yazdinejad2024robust}, as adversaries can refine their strategies by exploiting global model updates, compromising the learning process and model integrity.

To solve above challenges, one-shot FL, which restricts the communication rounds between clients and the server to a single iteration \cite{guha2019one}, has emerged as a promising solution. Existing one-shot FL methods can be broadly divided into three categories: 1) Knowledge distillation methods \cite{li2020practical, zhang2022dense, dai2024enhancing}. 2) Generative methods \cite{heinbaughdata, yang2024exploring, yang2023one, yang2024feddeo}. 3) Bayesian methods \cite{neiswanger2013asymptotically, jhunjhunwala2024fedfisher, hasan2024calibrated}. 
Knowledge distillation methods obtain global models by distilling knowledge from an ensemble of client models using auxiliary public data or synthetic data generated by the ensemble. However, these distillation steps impose a significant computational cost on the server and require careful hyperparameter tuning \cite{kurach2019large}. Generative methods aim to train the global model on the server by using synthetic samples that match the distribution of each client’s data. These approaches, though effective, brings about privacy concerns when the generated data closely mimics the original client data in the server \cite{rao2024privacy, carlini2023extracting}. Bayesian methods approximate each client's posterior distribution and aggregate them into a global model within a single communication round. However, these methods face high computational and memory costs, and using approximations to reduce these costs can lead to significant errors, compromising both calibration and accuracy \cite{neiswanger2013asymptotically, hasan2024calibrated}. Moreover, both of these three kinds of methods have poor performance and robustness when dealing with non-IID (Independent and Identically Distributed) data. 

Recently, FedPFT \cite{beitollahi2024parametric} has been proposed to leverage pre-trained models to improve both the accuracy and computation efficiency of one-shot FL. Specifically, each client uploads the Gaussian Mixture Models (GMMs) learned from class-conditional feature. Subsequently, the server trains a classifier using synthetic features generated from GMMs on the server. With competitive performance, low computation overhead and robustness to data heterogeneity, FedPFT \cite{beitollahi2024parametric} shows the feasibility of adapting the classifier with fixed pre-trained backbone in one-shot FL. However, GMMs may not effectively fit the local feature distributions. Additionally, different sampling and training strategies will influence the final performance. Actually, there is no need to generate clients' features for training the classifier. We can obtain the parameter-free classifier heads using feature statistics if we could get the global feature statistics. In this paper, we propose \textbf{FedCGS}, a novel \textbf{Fed}erated algorithm that \textbf{C}apture \textbf{G}lobal feature \textbf{S}tatistics in a communication-efficient and privacy-preserving way by leveraging pre-trained models. We then utilize these statistics to achieve global one-shot FL and personalized one-shot FL with competitive performance. Our key contributions are highlighted as follows:
\begin{itemize}
\item We propose FedCGS, a FL framework that utilizes pre-trained model to capture global feature statistics in computation-efficient and privacy-preserving manner.
\item Leveraging global feature statistics, we make use a parameter-free Naive Bayes classifier instead of the learnable linear classifier to achieve heterogeneity-resistant one-shot FL with competitive performance. Additionally, we propose personalized one-shot FL that each client uses these statistics as feature alignment regularizer for local training through one extra communication round to download the global statistics.
\item Extensive experiments show that FedCGS improve communication-accuracy frontier both in label shift and feature shift settings.
\end{itemize}

\section{Related Work}
\subsection{One-shot Federated Learning}
Existing one-shot FL methods can be broadly divided into three categories: 1) Knowledge distillation methods  \cite{li2020practical, zhang2022dense, dai2024enhancing}. 2) Generative methods \cite{heinbaughdata, yang2024exploring, yang2023one, yang2024feddeo}.  3) Bayesian methods \cite{neiswanger2013asymptotically, jhunjhunwala2024fedfisher, hasan2024calibrated}. 

Guha et al. \cite{guha2019one} introduce the first one-shot FL algorithm, which uses an ensemble of local models as the final global model. Knowledge distillation methods aim to reduce storage by consolidating the ensemble into a single model through knowledge distillation. FedKT \cite{li2020practical} achieve knowledge distillation using public data. DENSE \cite{zhang2022dense} eliminates the dependency on public data by using the ensemble of client-uploaded local models to train a generator, which then produces synthetic data for knowledge distillation. Similarly, Co-Boosting \cite{dai2024enhancing} progressively enhances both the ensemble model and the synthesized data to improve knowledge distillation. However, these methods demand substantial computation on the server and require meticulous hyperparameter tuning. 

Generative methods are proposed to train the global model on the server using generated data. DOSFL \cite{zhou2020distilled} trains the global model on the server using the distilled synthetic data from local clients. FedCVAE \cite{heinbaughdata} trains a conditional variation auto-encoder (CVAE) on each client, after which the decoders and the local label distribution are sent to the server to generate data for training a global model. FedDISC \cite{yang2024exploring} utilize data features to generate samples through auxiliary pre-trained diffusion model in context of semi-supervised FL. Like this, FGL \cite{zhang2023federated}, FedCADO \cite{yang2023one} and FedDEO \cite{yang2024feddeo} upload text prompts, classifiers and descriptions related with local data distribution respectively to provide suitable guidance for diffusion model. Nevertheless, generating data on the server that closely resembles client data raises privacy concerns \cite{carlini2023extracting}.

Bayesian methods estimate the posterior distribution of each client model within a Bayesian framework and aggregate them into a global model. \cite{neiswanger2013asymptotically} demonstrate how to aggregate local posteriors in a single communication round in FL. \cite{hasan2024calibrated} introduce a novel aggregation technique that interpolates between predictions from the Bayesian Committee Machine \cite{tresp2000bayesian}, reducing errors from approximations \cite{neiswanger2013asymptotically}. FedFisher \cite{jhunjhunwala2024fedfisher} and FedLPA \cite{liu2023fedlpa} utilize the empirical Fisher information matrix to approximate local posteriors, which are then used to compute the mode of the global posterior. Although Bayesian one-shot FL methods offer strong theoretical guarantees, they suffer from high computational costs, especially when dealing with large models.

Recently, FedPFT \cite{beitollahi2024parametric} was introduced to enhance both accuracy and computational efficiency in one-shot FL by leveraging pre-trained models. In this approach, each client uploads the GMMs learned from class-conditional features, and the server then trains a global classifier using synthetic features generated from these GMMs. 
FedPFT \cite{beitollahi2024parametric} demonstrates the effectiveness of adapting a classifier with a fixed pre-trained backbone in one-shot FL. However, sampling features to train a linear classifier can lead to a performance bottleneck, as GMMs may not effectively fit the local feature distributions, especially when local data is insufficient. Moreover, the variations of sampling strategies and classifier training configurations will affect the final performance. Distinct from this, we capture global feature statistics through a carefully designed uploading scheme and directly obtain the Gaussian Naive Bayes head as the global classifier in a training-free manner.

\subsection{Federated Learning with pre-trained Models}
Pre-trained models, such as ResNet \cite{resnet}, ViT \cite{Vit_ICLR2021} or BERT\cite{BERT_NAACL2019}, have been widely used to benefit downstream tasks. Recently, FL with pre-trained Models is becoming a popular topic with the increasing prevalence of pre-trained models \cite{nguyen2022begin, chen2022importance}. Due to the millions of parameters in pre-trained models, fine-tuning the entire model in FL leads to high communication costs and memory footprint issues. Therefore, recent research suggests that using parameter-efficient tuning methods, such as Adapter \cite{chen2024feddat}, Prompt Tuning \cite{guo2023pfedprompt, guo2023promptfl, li2024global, bai2024diprompt}, Low-Rank Adaption \cite{wu2024fedlora, nguyen2024flora}. In our method, we fix the backbone of the pre-trained model, and utilize it to extract features from local dataset for subsequent steps. 

\section{Method}
\label{sec:FedCGS}
\subsection{Problem Formulation}
Federated Learning (FL) involves a server coordinating with several clients to collaboratively train a global model without sharing private data. Suppose a FL system that consists of a central server and $M$ clients with their local datasets $D_1,\dots, D_M$,  correspondingly. These local datasets are sampled from $M$ distinct distributions and have different sizes. The goal of FL is to train a global  model $w$ that optimizes the following loss function across all clients:
\begin{equation}
\min_\mathbf{\theta}\mathcal{L}(\mathbf{\theta})=\frac{1}{M}\sum_{i\in[M]}\mathcal{L}_i(\mathbf{\theta}),
\end{equation}
where $\mathcal{L}_i(\mathbf{\theta}):=\frac{1}{|D_i|}\sum_{\delta_i\in D_i}\ell(\mathbf{\theta};\delta_i)$ is the empirical risk objective computed on the local data set $D_i$ at the $i$-th client, $\ell(\cdot,\cdot)$ is a loss function and $\delta_i$ denotes a sample from $D_i$. 

In contrast, the goal of personalized Federated Learning (pFL) is to personalize the global model with labeled data of each client to better adapt to the local data distribution:
\begin{equation}
\min_{\theta_{1:M}}\mathcal{L}(\theta_{1:M})=\frac{1}{M}\sum_{i\in[M]}\mathcal{L}_i( \theta_i).
\end{equation}
Here, $\theta_{1:M}$ denotes the collection of local models.

\subsection{Capture Global Feature Statistics}
\label{sec:Capture Global Feature Statistics}
As shown in Fig \ref{framework}, in our FedCGS, each client $i$ extracts class-specific features from its own dataset for each class $j\in\{1,\dots,C\}$ and calculates the statistics $S_i = \{B_i\} \cup \{(A_i^j, N_i^j) \mid j = 1, \ldots, C\}$ as follows:
% \begin{small}
\begin{align}
&A_i^j= \sum\nolimits_{x\in D_i^j}f(x), \\
&B_i=\sum\nolimits_{x\in D_i}f(x)^\top f(x), \\
&N_i^j=|D_i^j|,
\end{align}
% \end{small}
where $D_i^j=\{x_k\in D_i \mid y_k=j\}$ and $f$ denotes the feature extractor of the pre-trained model. After that, each client uploads $S_i$ to the server. Now, we demonstrate how to calculate global prototype and global empirical covariance (of features) with these uploaded parameters:
 
\begin{figure}[ht]
% \centering
\includegraphics[width=0.465\textwidth]{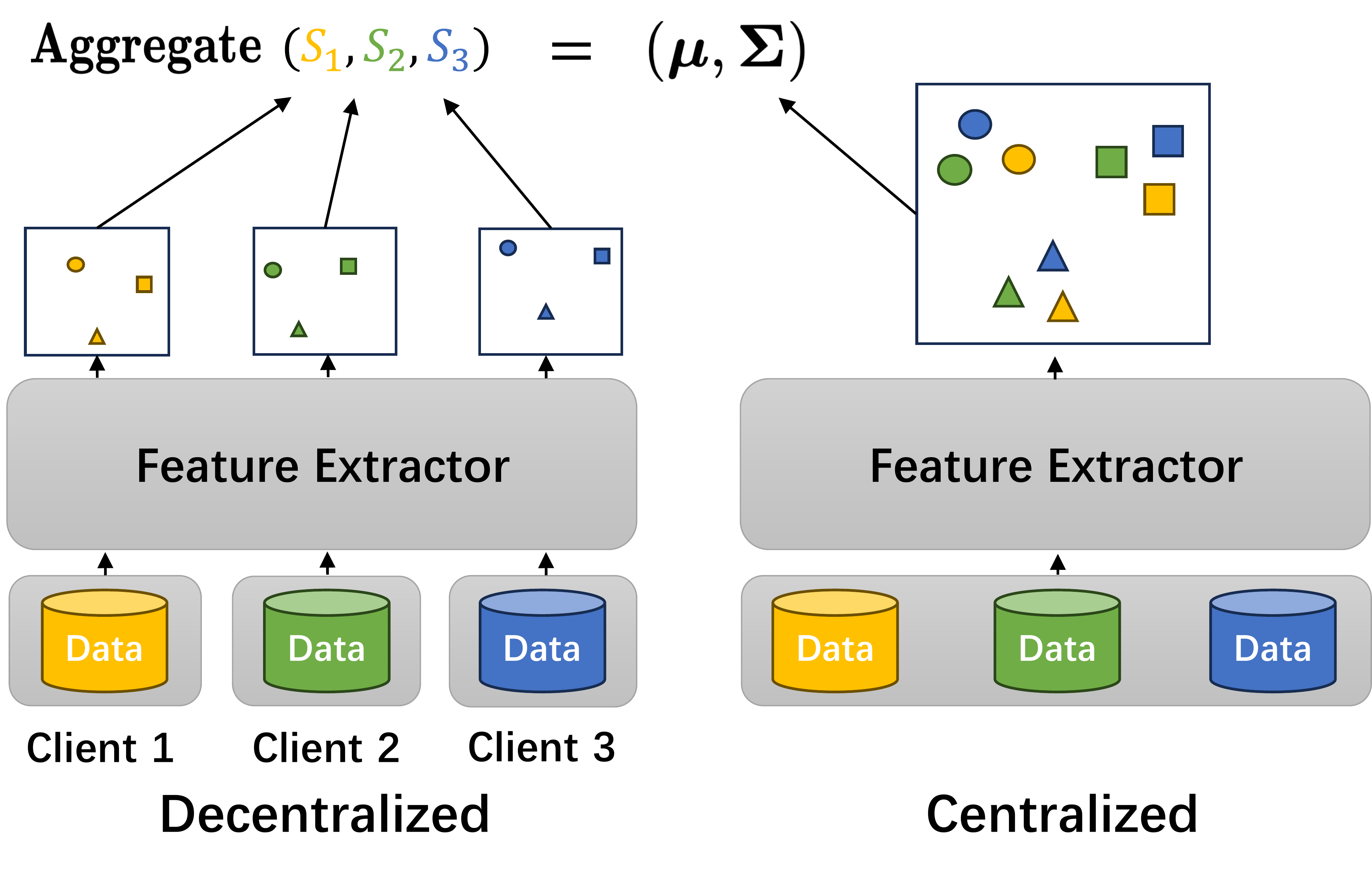}
\caption{Framework of our FedCGS. $S_i$ is the local statistics of client $i$ as shown in \ref{sec:Capture Global Feature Statistics}. We obtain global feature statistics through aggregating local feature statistics} \label{framework}
\end{figure} 

\noindent
\textbf{global prototype} $\boldsymbol{\mu}=[\mu^1,\mu^2,\cdots,\mu^C]$.\hspace{4pt} 
We denote $N^j$ \vspace{2pt} as the total number of instances belonging to class $j$ over all clients ($N^j=\sum_{i=1}^M N_{i}^j$) and $A^j$ as the sum of features belonging to class $j$ over all clients ($\small A^j=\sum_{i=1}^{M}A_i^j$):
\begin{small}
\begin{align} 
\mu^j&=\frac{1}{N^j}\sum_{i=1}^{M}\sum_{x\in D_i^j} f(x)=\frac{1}{N^j}\sum_{i=1}^{M}A_i^{j}=\frac{1}{N^j}A^{j},
\end{align}
\end{small}

\noindent
\textbf{global empirical covariance of features $\boldsymbol{\Sigma}$}. \hspace{4pt}  Here, we denote $A=\sum_{j=1}^{C}A^j$, $B=\sum_{i=1}^{M}B_i$, $N=\sum_{j=1}^C N^j$ and $f(x) \triangleq f$ for simplicity:
\begin{small}
\begin{align}
\boldsymbol{\Sigma}&=\frac{1}{N-1}\sum_{i=1}^M\sum_{x\in D_i}(f-\mu) ^\top(f-\mu) \notag \\
&=\frac{1}{N-1}\sum_{i=1}^M\sum_{x\in D_i}(f^\top f-\mu^\top f-f^\top\mu+\mu^\top\mu) \notag \\
&=\frac{1}{N-1}\sum_{i=1}^M(B_i-\mu^\top A_i-A_i^\top\mu+N_i\mu^\top\mu) \notag \\
&=\frac{1}{N-1}(B-\mu^\top A-A^\top\mu + N\mu^\top\mu),
\end{align}
\end{small}

\noindent
where $\mu$ denotes the global mean of features. We could get $\mu$ as follow:
\begin{small}
\begin{align} 
\mu&=\frac{1}{N}\sum_{i=1}^{M}\sum_{x\in D_i} f(x)=\frac{1}{N}\sum_{i=1}^{M}A_{i} = \frac{1}{N}A.
\end{align}
\end{small}

Algorithm \ref{algo:FedCGS} summarizes the procedure of FedCGS with pseudocode. Noticed that each client is required to upload the class-wise label count $N_i^j$, which could potentially lead to the leakage of label information. However, for subsequent calculations, only the aggregated label counts $N^j$ is necessary. Therefore, we can employ Secure Aggregation \cite{bonawitz2017practical} as shown in line 5, ensuring that the server does not gain access to any individual client’s label information. Detailed discussion about privacy can be found in Discussion.

Subsequently, we illustrate how to utilize global feature statistics to achieve global one-shot FL and personalized one-shot FL with outstanding performance.

\begin{algorithm}[t]
\small
\caption{Procedure of FedCGS}\label{algo:FedCGS}
\textbf{Server Executes:}
\begin{algorithmic}[1]

\FOR [\textbf{in parallel}]{client $i = 1$ to $M$} 
    \STATE $\{A_i^j, N_i^j\}_{j=1}^C$, $B_i \gets \textbf{ClientStats}(D_i)$
\ENDFOR

\FOR {each class $j \in C$} 
    \STATE $N^j \gets \mathbf{SecureAgg}\left(\sum_{j=1}^{M} N_i^j\right)$
    \STATE $A^j \gets \sum_{i=1}^{M} A_i^j$
    \STATE $\mu^j \gets \frac{1}{N^j} A^j$
\ENDFOR

\STATE $N \gets \sum_{j=1}^C N^j$ \\
\STATE $A \gets \sum_{i=1}^M \sum_{j=1}^C A_i^j$ \\
\STATE $B \gets \sum_{i=1}^M B_i$ \\
\STATE $\boldsymbol{\Sigma} \gets \frac{1}{N-1} \left(B - \mu^\top A - A^\top \mu + N \mu^\top \mu \right)$ \\

\end{algorithmic} 

\textbf{ClientStats}$(D_i)$
\begin{algorithmic}[1]
\FOR {each class $j \in C$} 
    \STATE Let $D_i^j = \{x_k \in D_i \mid y_k = j\}$, $|D_i^j| = N_i^j$ \\
    \STATE $A_i^j \gets \sum_{x \in D_i^j} f(x)$ \\  
\ENDFOR
\STATE $B_i \gets \sum_{x \in D_i} f(x)^\top f(x)$ \\
\RETURN $S_i = \{B_i\} \cup \{(A_i^j, N_i^j) \mid j = 1, \ldots, C\}$
\end{algorithmic}
\end{algorithm}

\subsection{FedCGS for global one-shot FL}
\label{sec:FedCGS_g}
Instead of sampling synthetic features from GMMs to directly train the linear head \cite{beitollahi2024parametric}, we fix the pre-trained models as backbone and use a Gaussian Naive Bayes classifier as the head of network, which can be configured directly from feature statistics. For class $j$, the class probability for a test point $x^{*}$ is:
\begin{align}
% \label{classifier}
p(&y^*=j|f(x^*),\boldsymbol{\pi},\boldsymbol{\mu},\boldsymbol{\Sigma})=\frac{\pi_j\mathcal{N}(f(x^*)|\mu^j,\boldsymbol{\Sigma}))}{\sum_{j^{\prime}}^C\pi_{j^{\prime}}\mathcal{N}(f(x^*)|\mu^{j^{\prime}},\boldsymbol{\Sigma})}\\
\label{classifier}
=&\frac{\pi_j\exp\left((\mu^{j})^{\top}\boldsymbol{\Sigma}^{-1}f(x^*)-\frac12(\mu^{j})^{\top}\boldsymbol{\Sigma}^{-1}\mu^j\right)}{\sum_{j^{\prime}}^C\pi_{j^{\prime}}\exp\left((\mu^{j^{\prime}})^\top\boldsymbol{\Sigma}^{-1}f(x^*)-\frac12(\mu^{j^{\prime}})^\top\boldsymbol{\Sigma}^{-1}\mu^{j^{\prime}}\right)},
\end{align}
where $\pi_{j}=\frac{N^j}{N}$. Therefore, the weight $W \in \mathbb{R}^{C\times d}$ and the bias $b\in \mathbb{R}^{C}$ for the classifier can be expressed as: 
\begin{align}
w_j=\boldsymbol{\Sigma}^{-1}\mu^j, b_j=\log\pi_j-\frac{1}{2}(\mu^j)^\top\boldsymbol{\Sigma}\mu^j.
\end{align}
Refer to \textbf{appendix} for detailed calculation.

\subsection{FedCGS for personalized one-shot FL}
Unlike existing one-shot FL, FedCGS can benefit personalized FL leveraging global feature statistics. The clients only need to execute one extra communication round to download global feature statistics after obtaining global feature statistics, which is the reason we call it personalized one-shot FL.

In personalized FL, each client usually owns insufficient data, making the locally learned feature representation prone to overfitting and poor generalization. Here, each client uses global prototypes $\boldsymbol{\mu}$ downloaded from server as a regularization term for better local feature representation learning (finetune entire model on local datasets). Specifically, taking client $i$ as an example, the objective of local training can be formulated as:
\begin{align}
\min_{\theta_{i}}\mathcal{L}(\theta_{i})&=\mathcal{L}_i(\theta_i)+ \lambda \mathcal{R}(\theta_i,\boldsymbol \mu),\notag \\ 
&=\mathcal{L}_i(\theta_i)+\frac{\lambda}{N_i^j} \sum_{j=1}^{C}\sum_{x\in D_i^j}\left\|f(x;\theta_i)-\mu^j\right\|_2^2.
\end{align}
Feature alignment regularizer $\mathcal{R(\cdot,\cdot)}$ has been proven to be highly effective in few-shot learning, domain adaption and federated learning \cite{li2020prototypical, tan2022fedproto, xu2023personalized}. Different from existing personalized FL methods that involves feature alignment, our global prototype $\boldsymbol{\mu}$ remains fixed while others, like FedProto \cite{tan2022fedproto}, updates it in the server with each communication round.

\section{Experiments}
To show the effectiveness of our proposed FedCGS, we conduct experiments on both the global one-shot FL and the personalized one-shot FL. More details and extra results are included in the \textbf{appendix}.

\begin{table*}[t]
\centering{
\begin{tabular}{@{}c|c|cccccc@{}}
\toprule
          & $\alpha$  & FedAvg (one-shot) & Ensemble & DENSE  & Co-Boosting  & FedPFT & FedCGS       \\
         \midrule
         & 0.05 & 13.97±1.26 & 38.81±0.89 & 31.26±0.73 & 44.37±0.42  & 56.08±0.49 & \textbf{63.95±0.00} \\
CIFAR10  & 0.1  & 27.28±1.34 & 57.29±0.54 & 56.21±0.24 & 60.41±0.67  & 56.43±0.23 & \textbf{63.95±0.00} \\
         & 0.5  & 51.91±0.46 & 66.00±0.42 & 62.42±0.43 & \textbf{67.43±0.36} & 56.80±0.18 & 63.95±0.00 \\
         \midrule
         & 0.05 & 17.57±0.58 & 22.43±0.63 & 14.31±0.61 & 20.30±0.76  & 36.79±0.21 & \textbf{39.95±0.00} \\
CIFAR100 & 0.1  & 21.46±0.38 & 28.07±0.47 & 17.21±0.36 & 24.63±0.64  & 37.16±0.34 & \textbf{39.95±0.00} \\
         & 0.5  & 35.26±0.22 & 37.89±0.49 & 26.49±0.32 & 34.43±0.45  & 37.95±0.27 & \textbf{39.95±0.00} \\
        \midrule
         & 0.05 & 16.75±0.63 & 42.26±0.54 & 37.49±0.42 & 41.90±0.38  & 42.55±0.24  & \textbf{57.77±0.00} \\
SVHN     & 0.1  & 24.88±0.39 & 53.34±0.23 & 51.53±0.37 & 57.13±0.28  & 43.03±0.17  & \textbf{57.77±0.00} \\
         & 0.5  & 44.39±0.46 & 82.93±0.29 & 77.44±0.26 & \textbf{84.65±0.24}  & 43.84±0.42 & 57.77±0.00 \\
         \bottomrule
\end{tabular}
  \caption{Test accuracy (\%) of the global model for different methods over three datasets at three levels of statistical heterogeneity (where a lower $\alpha$ signifies more heterogeneity).} 
    \label{tab:gfl}
}
\end{table*}

\begin{table}[]
\centering
\begin{tabular}{c|c|cccc}
\toprule
                            & Domain & FedPFT & FedCGS     \\
\midrule                    
\multirow{5}{*}{PACS}       & P    & 87.98±1.13 & \textbf{91.02±0.00} \\
                            & A    & 58.34±2.30 & \textbf{64.21±0.00} \\
                            & C    & 46.15±1.55 & \textbf{52.30±0.00} \\
                            & S    & 36.96±1.62 & \textbf{41.03±0.00} \\
                            \cmidrule{2-4}
                            & Avg. & 57.36±1.65  & \textbf{62.14±0.00} \\
                            \midrule
\multirow{5}{*}{OfficeHome} & P    & 67.98±0.71 & \textbf{68.75±0.00} \\
                            & A    & 55.74±0.46 & \textbf{57.44±0.00} \\
                            & C    & 39.26±1.01 & \textbf{40.99±0.00} \\
                            & R    & 67.75±0.68 & \textbf{73.28±0.00} \\
                            \cmidrule{2-4}
                            & Avg. & 57.68±0.72 & \textbf{60.12±0.00} \\
                             \bottomrule
\end{tabular}
 \caption{Test accuracy (\%)  of the global model of different methods in PACS and OfficeHome dataset on the domain generalization test mechanism.} 
 % \label{-20pt}
    \label{tab:gfl_f}
\end{table}

\subsection{Global one-shot FL}
\textbf{Datasets and Partitions.} Our experiments are conducted for classification task on the following image datasets: SVHN \cite{netzer2011reading}, CIFAR10 \cite{krizhevsky2009cifar}, CIFAR100 \cite{kri2009cifar100}, PACS \cite{li2017deeper}, and OfficeHome \cite{venkateswara2017deep}. For label shift scenario, we use Dirichlet distribution to generate disjoint non-IID client training datasets as same as other global one-shot FL methods \cite{zhang2022dense, heinbaughdata, dai2024enhancing} for fair comparison. For feature shift scenario, we follow the domain generalization settings in \cite{bai2024diprompt}. Specifically, 
we select three domains for training and distribute their data across $M$ clients. Data from a single domain may be spread across multiple clients, but each client belongs to only one domain. The global model is then tested on the target domain. For all datasets, we adopt a commonly used ResNet18 \cite{resnet} pre-trained on ImageNet as the backbone.

\noindent
\textbf{Baselines.} To evaluate our proposed method, we compare it with the canonical baseline FedAvg \cite{McMahan17}, Ensemble and other state-of-the-art (SOTA) methods in one-shot FL : Dense \cite{zhang2022dense}, Co-Boosting \cite{dai2024enhancing}, and FedPFT \cite{beitollahi2024parametric}. DENSE is the first data-free knowledge distillation method for one-shot federated learning, which distills knowledge from an ensemble of client models. Ensemble maintains an ensemble of local models at the server, serving as the upper bound for DENSE. Ensemble is a strong baseline, but it requires storing all local models on the server, which leads to significant storage overhead and limits scalability and efficiency. Co-Boosting is the SOTA one-shot FL method based on data-free knowledge distillation, which improves the quality of the ensemble model and data generation simultaneously. We avoid comparing with baselines that are inherently multi-round, such as FedProx \cite{sahu2019convergence}, as their performance would be similar to FedAvg after a single round. Additionally, we exclude algorithms that need auxiliary data \cite{lin2020ensemble} or pre-trained generative models \cite{heinbaughdata, yang2024feddeo, yang2024exploring, yang2023one} to ensure fairness of comparison. In feature shift scenario, since the official implementations for this settings are not provided, we mainly compare our FedCGS with FedPFT. Additional comparisons with other baselines can be found in \textbf{appendix}.

\noindent
\textbf{Configurations.} In the label shift scenario, we simulate a federated learning environment with 10 clients and set the client participation ratio $\rho$ to 1, as in existing one-shot federated learning methods \cite{zhang2022dense, dai2024enhancing}. In the feature shift scenario, the data from each source domain is randomly split across 5 clients, resulting in 15 clients in total for 3 domains. For methods involving backpropagation training (FedAvg, Ensemble, DENSE, Co-Boosting, FedPFT), we set the batch size to 128, the number of epochs to 50, and use the Stochastic Gradient Descent (SGD) optimizer with momentum = 0.9 and the learning rate = 0.01. For data-free knowledge distillation, we use the same generator as in \citep{dai2024enhancing, zhang2022dense}. It is trained with the Adam optimizer, a learning rate of 1e-3, for 30 epochs. FedPFT uploads the GMMs $\mathcal{G}(K_g)$ with a diagonal covariance. Here $\mathcal{G}(K_g)$ is the family of all Gaussian mixture distributions comprised of $K_g$ components. Specifically, we set $K_g$ as 10. For each client, the class-wise label counts are used to sample the corresponding number of features from each GMMs to construct the training dataset.

\noindent
\textbf{Experimental Results.} We verify that the proposed FedCGS outperforms existing global one-shot FL methods in most cases. As shown in Table \ref{tab:gfl} and Table \ref{tab:gfl_f}, we test the performance of global one-shot method in the label-shift setting and feature-shift setting, respectively. For label shift setting, we have the following observations: (1) Traditional FL algorithm FedAvg, which usually requires multiple rounds to converge, performs poorly in this one-shot scenario. (2) Ensemble is the upper bound of DENSE, and it is surpassed by Co-Boosting in some cases. Co-Boosting introduces several strategies to improve both the generated data and the quality of the ensemble model. However, these improvements cannot be achieved under high statistical heterogeneity. (3) All baselines except FedPFT suffer from a large performance deterioration under high statistical heterogeneity ($\alpha=0.1,0.05$) (4) Most existing methods exhibit considerable performance variability with different random seeds. (5) In our FedCGS, dataset partitioning does not affect the global quantities $A$ and $B$, so our FedCGS is immune to statistical heterogeneity and achieves competitive results compared to most existing global one-shot FL methods, especially under extremely high statistical heterogeneity ($\alpha=0.05$). Specifically, FedCGS surpasses the best baseline by substantial margins with 7.87\%, 3.16\%, 15.22\% on CIFAR10, CIFAR100, SVHN respectively. Additionally, since no training is involved, the performance remains stable. For feature shift, as demonstrated in our results, FedCGS achieves the highest average accuracy and consistently outperforms FedPFT across all target domains. This further substantiates the efficacy of our proposed method.

\subsection{Personalized one-shot FL}
\textbf{Datasets and Partitions.} As same as global one-shot FL, we conduct experiments on SVHN \cite{netzer2011reading}, CIFAR10 \cite{krizhevsky2009cifar}, CIFAR100 \cite{kri2009cifar100} using ResNet18 pre-trained on ImageNet. For data partitions, we follow the previous personalized FL methods \cite{zhang2020personalized,huang2021personalized,xu2023personalized} that all clients have same data size, owning $s$\% of data (20\% by default) uniformly sampled from all classes and $(100-s)$\% from a set of dominant classes. 

\noindent
\textbf{Baselines.} We compare the performance of our method against following baselines: Local-only, where each client trains its model locally; FedAvg \cite{McMahan17} with vanilla local training; FedAvg-FT that learns a single global model and locally fine-tuned on local datasets, a simple but strong baseline; FedProto \cite{tan2022fedproto}, FL with only prototype sharing.

\noindent
\textbf{Configurations.} We use ResNet18 pre-trained on ImageNet for all datasets. During local training phase for each client, we employ mini-batch SGD as the local optimizer and set the batch size to 128, the local epoch to 1 for traditional personalized FL, 200 for Local-only and ours. The momentum is set to 0.5, the learning rate is set to 0.01, the weight decay is set to 5e-4 as \cite{xu2023personalized}. The number of global communication rounds for traditional personalized FL is set to 100 across all datasets. All results are reported averaged across 3 random seeds.

\noindent
\textbf{Experimental Results.} As shown in Table \ref{tab:pfl}, FedProto \cite{tan2022fedproto} and ours consistently outperform the baseline FedAvg \cite{McMahan17} and Local-only. This suggests that the discrepancy between local and global feature distributions could lead to higher generalization error and the feature alignment regularization is a good solution. Additionally, We observer that our FedCGS performs similarly to, and in some cases better than, FedProto \cite{tan2022fedproto} (e.g., on CIFAR10). That means that fixed global prototypes in FedCGS plays same role as the updated global prototypes in FedProto \cite{tan2022fedproto}, which illustrates that features extracted by pre-trained models are sufficiently strong enough to capture meaningful patterns \cite{janson2022simple, beitollahi2024parametric}. FedAvg-FT achieves best results across all datasets, which verifies the statement that FedAvg-FT is a strong baseline, even outperforming many personalized FL approaches. We want to emphasize that the contribution of our personalized one-shot FL lies in providing a method to enhance local model performance without requiring multiple communication rounds.

\begin{table}[ht]
\centering
\begin{tabular}{l|c|c|c}

           \toprule 
           & CIFAR10    & CIFAR100   & SVHN         \\
           \midrule
Local-only & 76.25±0.26 & 54.06±0.28 & 80.73±0.37   \\
FedAvg     & 75.70±0.58 & 48.17±0.51 & 81.16±0.41  \\
FedAvg-FT  & 80.47±0.32 & 60.97±0.24 & 84.03±0.23   \\
FedProto   & 78.07±0.76 & 55.97±0.72 & 82.14±0.57   \\
FedCGS     & 78.30±0.39 & 55.56±0.61 & 81.20±0.45 \\
\bottomrule
\end{tabular}
  \caption{The comparison of final test accuracy (\%) on different datasets.} 
    \label{tab:pfl}
    \vspace{-10pt}
\end{table}

\subsection{In-deepth Study}
\noindent
\textbf{Comparing with centralized feature statistics.} Here we measure the deviation between global feature statistics $(\boldsymbol{\mu}, \boldsymbol{\Sigma})$ captured by ours and the ground truth $(\boldsymbol{\hat{\mu}}, \boldsymbol{\hat{\Sigma}})$ using $L_2$ norms ($\Delta\boldsymbol{\mu}=\|\boldsymbol{\mu}-\boldsymbol{\hat{\mu}}\|_2, \Delta\boldsymbol{\Sigma}=\|\boldsymbol{\Sigma}-\boldsymbol{\hat{\Sigma}}\|_2$). We conduct experiments on CIFAR10 using ResNet18 as the pre-trained model. As shown in Table \ref{tab:deviation}, 
the deviations remain between $10^{-7}\sim10^{-5}$ and are not sensitive to the number of clients $M$ or the degree of heterogeneity $\alpha$. This level of error does not affect the final classifier's configuration, and the accuracy evaluated in test dataset remains consistent across different scenarios. Same conclusion could be obtained on other datasets, refer to \textbf{appendix} for details.

\begin{table}[]
% \centering{
% \small
% \scriptsize
\fontsize{9pt}{10pt}\selectfont % 将字体设置为 8pt，行间距为 10pt
\setlength{\tabcolsep}{4pt} % 将列间距设置为 12pt
\begin{tabular}{@{}c|ccc|ccc@{}}

\toprule
     &  \multicolumn{3}{c}{M=10}  & \multicolumn{3}{c}{M=50}  \\
     \midrule
     $\alpha$ & $\Delta\boldsymbol{\mu}$ & $\Delta\boldsymbol{\Sigma}$     & Acc & $\Delta\boldsymbol{\mu}$ & $\Delta\boldsymbol{\Sigma}$ & Acc     \\
     \midrule
0.05 & 2.18E-06  & 1.47E-05   &63.95& 7.90E-07     & 7.83E-06 & 63.95  \\
0.1  & 2.41E-06     & 1.64E-05   &63.95& 6.32E-07     & 6.66E-06 & 63.95  \\
0.5  & 2.46E-06     & 1.61E-05   &63.95& 1.68E-07     & 4.37E-06 & 63.95  \\
\bottomrule
\end{tabular}
  \caption{The $L_2$ error between the FedCGS output and the real global feature statistics when using pre-trained ResNet18 on CIFAR10 (average 3 runs)} 
  \vspace{-10pt}
    \label{tab:deviation}
\end{table}

\begin{figure}[ht]
\includegraphics[width=0.45\textwidth]{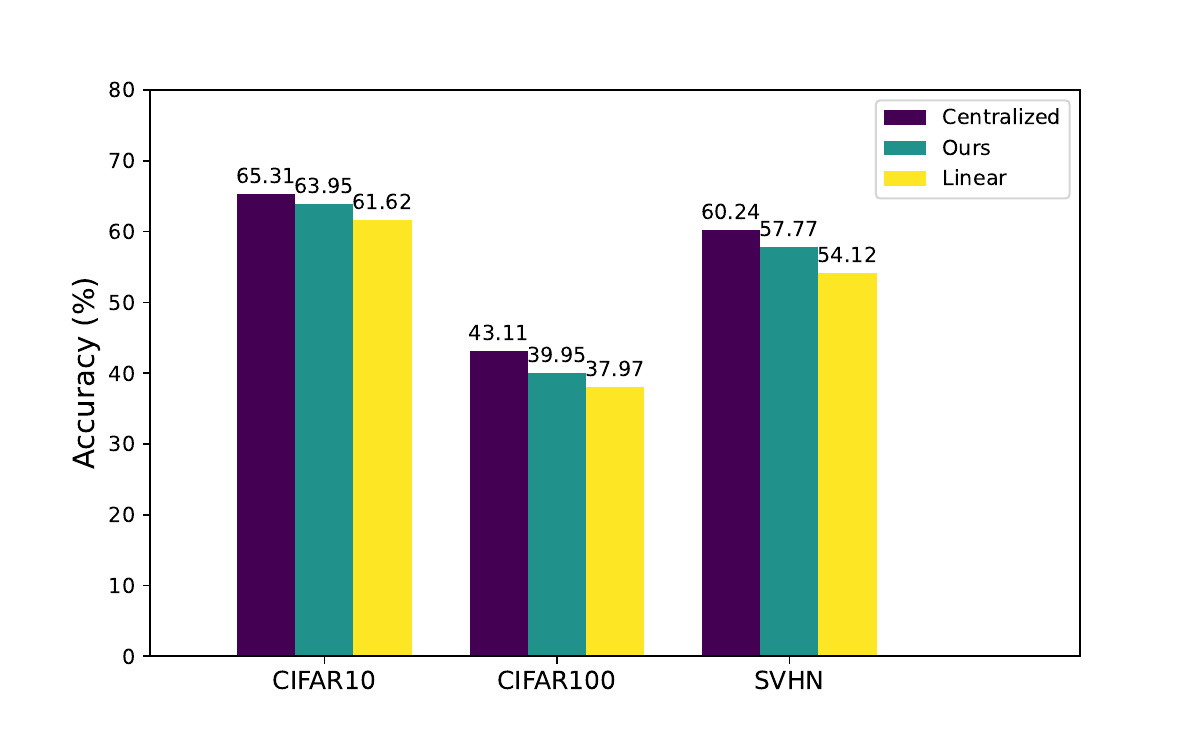}
\caption{Performance comparison using different classifier configurations.} \label{head}
\end{figure}

\vspace{5pt}
\noindent
\textbf{Naive Gaussian classifier head.} \label{head_analysis} We use the real global feature statistics captured by FedCGS to generate features in a GMM style and train the linear head as FedPFT. Additionally, we train a linear head in the centralized scenario (using all raw features) and compare the results with our Naive Gaussian classifier head. As shown in Fig \ref{head}, the Naive Gaussian classifier head achieves performance comparable to "Centralized" and outperforms "Linear" (the linear head trained with generated features). "Linear" can be regarded as the upper bound of FedPFT, since it uses global feature statistics to generate features for training, while FedPFT uses local statistics. This explains why FedCGS outperforms FedPFT.

\begin{figure}[ht]
\includegraphics[width=0.45\textwidth]{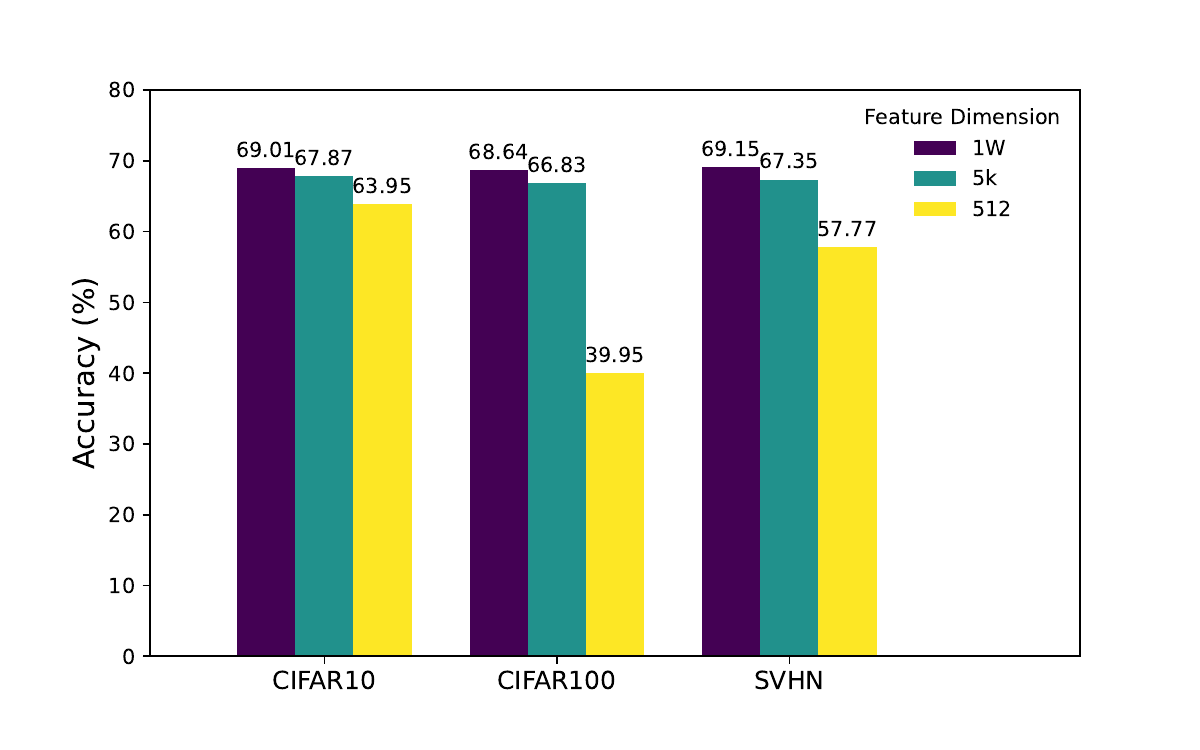}
\caption{Results of Feature expansion.} \label{fe}
\end{figure}

\noindent
\textbf{Feature expansion.} Given that pre-trained feature extractors may not expressive enough for clear class separation, we inject a same random projection layer with nonlinear activation between each client's pre-trained feature representations and output to enhance linear separability. We conduct experiments on three datasets using pre-trained ResNet18. As shown in Fig \ref{fe}, feature expansion improves performance across all datasets, with particularly notable gains on the more challenging CIFAR100 dataset. Feature expansion significantly enhances performance but also increases communication overhead (larger $d$), necessitating a trade-off between the performance and communication overhead. 

\noindent
\textbf{Different pre-trained models.} Our method relies on the feature extraction capabilities of pre-trained models, so here we conducted experiments on three datasets using different pre-trained models including ResNet18, ResNet50, MobileNetV2, EfficientNetB0. As shown in Table \ref{tab:arch}, using pre-trained ResNet50 consistently outperforms others. This indicates that our method could achieve significantly great performance if we own a strong enough pre-trained model.

\begin{table}[ht]
\centering
\begin{tabular}{l|c|c|c}
           \toprule 
           & CIFAR10   & CIFAR100   & SVHN \\
           \midrule
ResNet18       & 63.95   & 39.95    & 57.77 \\
ResNet50       & 70.32   & 46.08    & 60.31 \\
MobilenetV2    & 57.87   & 24.39    & 37.49 \\
EfficientNetB0 & 56.80    & 31.20     & 53.80  \\
\bottomrule
\end{tabular}
  \caption{Evaluation (\%) of FedCGS using different pre-trained model.} 
    \label{tab:arch}
\end{table}

\noindent
\textbf{Comparison with CCVR.}
We note that CCVR \cite{luo2021no} shares similar idea as our proposed FedCGS. Specifically, CCVR uploads all clients' local label distribution, class-wise mean and class-wise covariance, and utilizing them to compute the global features distribution statistics in the server. Finally, virtual features are generated using GMMs to retrain the classifier as the final global classifier. However, there are several key differences which we would like to mention. Firstly, in CCVR, each client uploads the class-wise covariance, which causes significant communication overhead. Secondly, the computation process for global statistics in CCVR is incompatible with Secure Aggregation, like ours. Thirdly, CCVR faces a challenge similar to FedPFT \cite{beitollahi2024parametric}, where variations in sampling strategies and training configurations affect the results.

\section{Discussion}
\label{discussion}
\subsection{Communication Overhead}
Our proposed method only transmits local feature statistics $S_i$, rather than model updates, to the server. Taking the case that using pre-trained ResNet18 on CIFAR10 as an example, the transmitted parameters of FedAvg/DENSE/Co-Boosting is $|\theta|=$ 11,181,642, FedPFT is $(2d+1)K_gC=$ 102,500 and ours is $(C+d)\times d+C=$ 267,274. Our proposed FedCGS achieves higher and more stable accuracy while keeping communication overhead low.

\subsection{Privacy Discussion}
Within the framework of the FedCGS algorithm, each client $i$ uploads statistics $S_i$ as shown in Algorithm \ref{algo:FedCGS}. Uploading these statistics may arise concerns of potential privacy risk. However, FedCGS only utilizes the aggregated values $\small N^j$,$A^j$,$B$ rather than individual values $\small N^j_i,A^j_i,B_i$ related with a specific client $i$, therefore Secure Aggregation \cite{bonawitz2017practical, mai2024rflpa} can be employed to ensure that the server only receive the aggregated values. Existing one-shot FL methods \cite{zhang2022dense, dai2024enhancing, heinbaughdata, jhunjhunwala2024fedfisher, yang2024exploring, yang2023one, yang2024feddeo, beitollahi2024parametric} all need to utilize the individual values from each client for subsequent steps. For example, Co-Boosting requires the parameters of each client model to build the ensemble, while FedPFT needs the class information from each client for sampling. This means that Secure Aggregation cannot be employed to enhance their privacy-preserving capability. 

\begin{figure}[h]
 \centering
\includegraphics[width=1\linewidth]{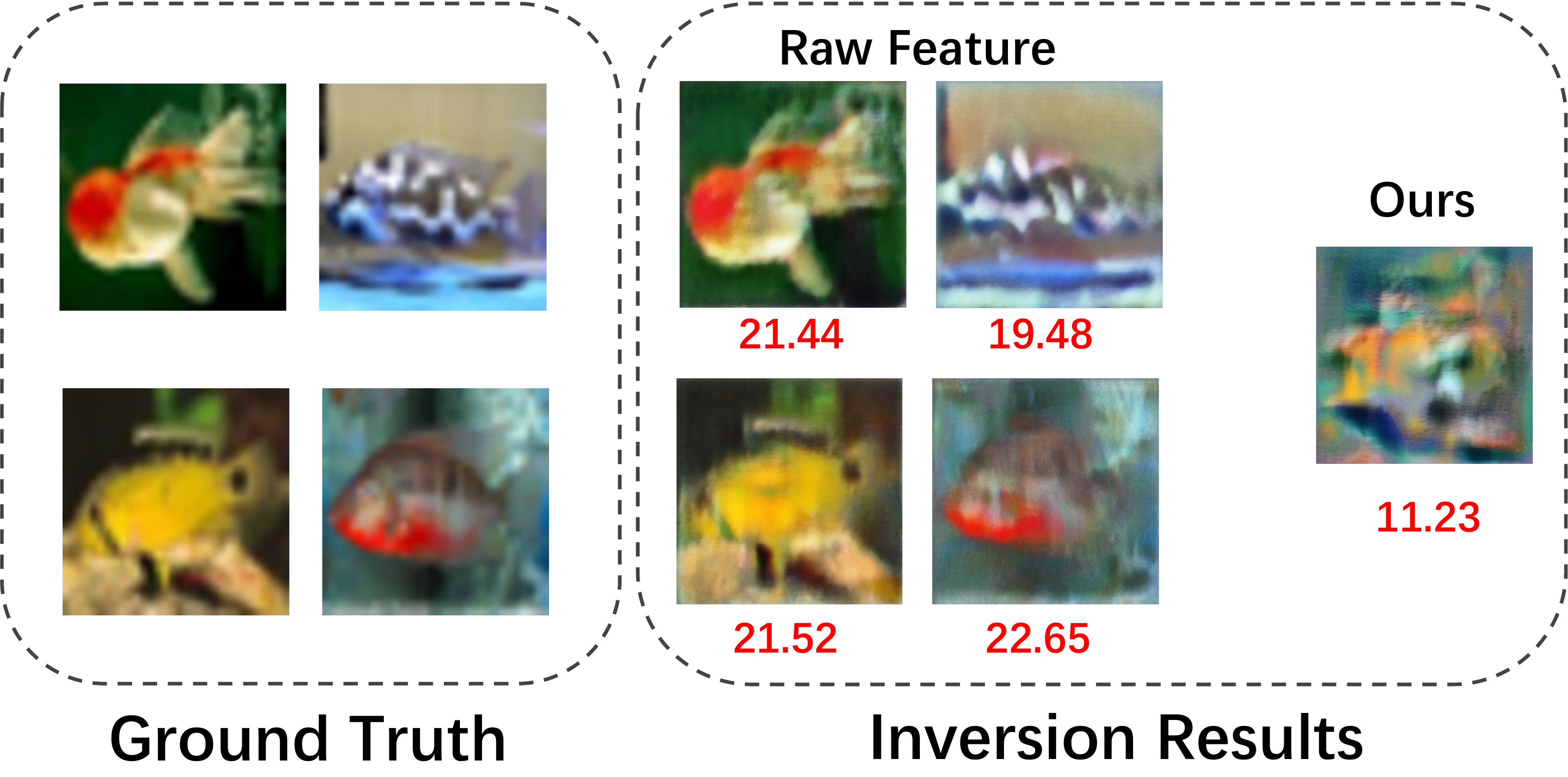}
\caption{Results of inversion attacks on CIFAR100. Assume a client has only 4 "aquarium fish" samples, as shown in \textbf{Ground Truth}. The server attempts to reconstruct a specific data sample from this client. If the server has access to the \textbf{Raw Feature} of each sample, the reconstructed results are clear. However, when using our uploaded variables, the results are poor. The PSNR value (red) is displayed below each reconstructed image as a quantitative measure.}
\label{fig:inversion}
\end{figure}

Considered that applying Secure Aggregation may incur additional communication overhead. We offer two options: one is to perform Secure Aggregation for all the local uploaded variables, and the other is to apply it only to client label counts $\small\{N_i^j\}_{j=1}^C$ to prevent leakage of the local label distribution, as shown in Algorithm \ref{algo:FedCGS}. For $\small \{A_i^j\}_{j=1}^C$ and $\small B_i$, it's difficult to recover private data of client $i$ from these variables, as they represent the sum of many features. To empirically analyze it, we utilize the feature inversion technology \cite{inversion2}. Consider a favorable scenario for the attacker, where client $j$ has only a few samples from the same class. In this case, the server, acting as an attacker, attempts to reconstruct a specific data sample from client $j$ using the uploaded variables. The results in Figure \ref{fig:inversion} indicate that the uploaded variables cannot be used to recover the client's original data.

\subsection{Limitations}
Our work shares some limitations with existing one-shot FL methods. Firstly, FedCGS is primarily designed for image classification tasks, the adaptation of this framework to more complex tasks, such as generation, remains underexplored. Secondly, the performance of FedCGS depends on the pretrained model's ability to extract meaningful features. Addressing these limitations could significantly improve the applicability and effectiveness of this approach in real-world FL scenarios.

\section{Conclusion}
We introduce FedCGS, a novel FL framework that leverages pre-trained models to capture global feature statistics. By utilizing these global feature statistics, we employ a parameter-free Naive Bayes classifier instead of a learnable linear classifier, enabling heterogeneity-resistant one-shot FL with competitive performance. Additionally, we propose a personalized one-shot FL approach, where each client uses these statistics as a feature alignment regularizer for local training, facilitated by one additional communication round to download the global statistics. Extensive experiments demonstrate that FedCGS enhances the communication-accuracy trade-off in various scenarios.

\section{Acknowledgements}
This work was supported by the Chinese Academy of Sciences under grant No. XDB0690302.
\appendix
% \nobibliography*
\bibliography{aaai25}

\begin{thebibliography}{55}
\providecommand{\natexlab}[1]{#1}

\bibitem[{Bai et~al.(2024)Bai, Zhang, Guo, Li, Guo, Hou, Han, and Lu}]{bai2024diprompt}
Bai, S.; Zhang, J.; Guo, S.; Li, S.; Guo, J.; Hou, J.; Han, T.; and Lu, X. 2024.
\newblock DiPrompT: Disentangled Prompt Tuning for Multiple Latent Domain Generalization in Federated Learning.
\newblock In \emph{Proceedings of the IEEE/CVF Conference on Computer Vision and Pattern Recognition}, 27284--27293.

\bibitem[{Beitollahi et~al.(2024)Beitollahi, Bie, Hemati, Brunswic, Li, Chen, and Zhang}]{beitollahi2024parametric}
Beitollahi, M.; Bie, A.; Hemati, S.; Brunswic, L.~M.; Li, X.; Chen, X.; and Zhang, G. 2024.
\newblock Parametric Feature Transfer: One-shot Federated Learning with Foundation Models.
\newblock \emph{arXiv preprint arXiv:2402.01862}.

\bibitem[{Bonawitz et~al.(2017)Bonawitz, Ivanov, Kreuter, Marcedone, McMahan, Patel, Ramage, Segal, and Seth}]{bonawitz2017practical}
Bonawitz, K.; Ivanov, V.; Kreuter, B.; Marcedone, A.; McMahan, H.~B.; Patel, S.; Ramage, D.; Segal, A.; and Seth, K. 2017.
\newblock Practical secure aggregation for privacy-preserving machine learning.
\newblock In \emph{proceedings of the 2017 ACM SIGSAC Conference on Computer and Communications Security}, 1175--1191.

\bibitem[{Carlini et~al.(2023)Carlini, Hayes, Nasr, Jagielski, Sehwag, Tramer, Balle, Ippolito, and Wallace}]{carlini2023extracting}
Carlini, N.; Hayes, J.; Nasr, M.; Jagielski, M.; Sehwag, V.; Tramer, F.; Balle, B.; Ippolito, D.; and Wallace, E. 2023.
\newblock Extracting training data from diffusion models.
\newblock In \emph{32nd USENIX Security Symposium (USENIX Security 23)}, 5253--5270.

\bibitem[{Chen et~al.(2024)Chen, Zhang, Krompass, Gu, and Tresp}]{chen2024feddat}
Chen, H.; Zhang, Y.; Krompass, D.; Gu, J.; and Tresp, V. 2024.
\newblock Feddat: An approach for foundation model finetuning in multi-modal heterogeneous federated learning.
\newblock In \emph{Proceedings of the AAAI Conference on Artificial Intelligence}, volume~38, 11285--11293.

\bibitem[{Chen et~al.(2022)Chen, Tu, Li, Shen, and Chao}]{chen2022importance}
Chen, H.-Y.; Tu, C.-H.; Li, Z.; Shen, H.-W.; and Chao, W.-L. 2022.
\newblock On the importance and applicability of pre-training for federated learning.
\newblock \emph{arXiv preprint arXiv:2206.11488}.

\bibitem[{Chen et~al.(2023)Chen, Wan, Prakash, Zhang, Yuan, Gong, Fu, and Pan}]{chen2023workie}
Chen, R.; Wan, Q.; Prakash, P.; Zhang, L.; Yuan, X.; Gong, Y.; Fu, X.; and Pan, M. 2023.
\newblock Workie-talkie: accelerating federated learning by overlapping computing and communications via contrastive regularization.
\newblock In \emph{Proceedings of the IEEE/CVF international conference on computer vision}, 16999--17009.

\bibitem[{Dai et~al.(2024)Dai, Zhang, Li, Liu, Yang, and Han}]{dai2024enhancing}
Dai, R.; Zhang, Y.; Li, A.; Liu, T.; Yang, X.; and Han, B. 2024.
\newblock Enhancing One-Shot Federated Learning Through Data and Ensemble Co-Boosting.
\newblock \emph{arXiv preprint arXiv:2402.15070}.

\bibitem[{Devlin et~al.(2019)Devlin, Chang, Lee, and Toutanova}]{BERT_NAACL2019}
Devlin, J.; Chang, M.; Lee, K.; and Toutanova, K. 2019.
\newblock {BERT:} Pre-training of Deep Bidirectional Transformers for Language Understanding.
\newblock In \emph{NAACL}, 4171--4186. Association for Computational Linguistics.

\bibitem[{Dosovitskiy et~al.(2021)Dosovitskiy, Beyer, Kolesnikov, Weissenborn, Zhai, Unterthiner, Dehghani, Minderer, Heigold, Gelly, Uszkoreit, and Houlsby}]{Vit_ICLR2021}
Dosovitskiy, A.; Beyer, L.; Kolesnikov, A.; Weissenborn, D.; Zhai, X.; Unterthiner, T.; Dehghani, M.; Minderer, M.; Heigold, G.; Gelly, S.; Uszkoreit, J.; and Houlsby, N. 2021.
\newblock An Image is Worth 16x16 Words: Transformers for Image Recognition at Scale.
\newblock \emph{ICLR}.

\bibitem[{Guan et~al.(2024)Guan, Zhou, Gu, and Li}]{guan2024gie}
Guan, Z.; Zhou, Y.; Gu, X.; and Li, B. 2024.
\newblock GIE: Gradient Inversion with Embeddings.
\newblock In \emph{2024 IEEE International Conference on Multimedia and Expo (ICME)}, 1--6. IEEE Computer Society.

\bibitem[{Guha, Talwalkar, and Smith(2019)}]{guha2019one}
Guha, N.; Talwalkar, A.; and Smith, V. 2019.
\newblock One-shot federated learning.
\newblock \emph{arXiv preprint arXiv:1902.11175}.

\bibitem[{Guo, Guo, and Wang(2023)}]{guo2023pfedprompt}
Guo, T.; Guo, S.; and Wang, J. 2023.
\newblock {pFedPrompt: Learning Personalized Prompt for Vision-Language Models in Federated Learning}.
\newblock In \emph{Proceedings of the ACM Web Conference 2023}, 1364--1374.

\bibitem[{Guo et~al.(2023)Guo, Guo, Wang, Tang, and Xu}]{guo2023promptfl}
Guo, T.; Guo, S.; Wang, J.; Tang, X.; and Xu, W. 2023.
\newblock {Prompt{FL}: Let federated participants cooperatively learn prompts instead of models-federated learning in age of foundation model}.
\newblock \emph{IEEE Transactions on Mobile Computing}.

\bibitem[{Hasan et~al.(2024)Hasan, Zhang, Guo, Chen, and Poupart}]{hasan2024calibrated}
Hasan, M.; Zhang, G.; Guo, K.; Chen, X.; and Poupart, P. 2024.
\newblock Calibrated one round federated learning with bayesian inference in the predictive space.
\newblock In \emph{Proceedings of the AAAI conference on artificial intelligence}, volume~38, 12313--12321.

\bibitem[{He et~al.(2016)He, Zhang, Ren, and Sun}]{resnet}
He, K.; Zhang, X.; Ren, S.; and Sun, J. 2016.
\newblock Deep residual learning for image recognition.
\newblock In \emph{CVPR}, 770--778.

\bibitem[{Heinbaugh, Luz-Ricca, and Shao(2023)}]{heinbaughdata}
Heinbaugh, C.~E.; Luz-Ricca, E.; and Shao, H. 2023.
\newblock Data-Free One-Shot Federated Learning Under Very High Statistical Heterogeneity.
\newblock In \emph{The Eleventh International Conference on Learning Representations}.

\bibitem[{Huang et~al.(2021)Huang, Chu, Zhou, Wang, Liu, Pei, and Zhang}]{huang2021personalized}
Huang, Y.; Chu, L.; Zhou, Z.; Wang, L.; Liu, J.; Pei, J.; and Zhang, Y. 2021.
\newblock Personalized cross-silo federated learning on non-iid data.
\newblock In \emph{Proceedings of the AAAI conference on artificial intelligence}, volume~35, 7865--7873.

\bibitem[{Janson et~al.(2022)Janson, Zhang, Aljundi, and Elhoseiny}]{janson2022simple}
Janson, P.; Zhang, W.; Aljundi, R.; and Elhoseiny, M. 2022.
\newblock A simple baseline that questions the use of pretrained-models in continual learning.
\newblock \emph{arXiv preprint arXiv:2210.04428}.

\bibitem[{Jhunjhunwala, Wang, and Joshi(2024)}]{jhunjhunwala2024fedfisher}
Jhunjhunwala, D.; Wang, S.; and Joshi, G. 2024.
\newblock FedFisher: Leveraging Fisher Information for One-Shot Federated Learning.
\newblock In \emph{International Conference on Artificial Intelligence and Statistics}, 1612--1620. PMLR.

\bibitem[{Kairouz et~al.(2021)Kairouz, McMahan, Avent, Bellet, Bennis, Bhagoji, Bonawitz, Charles, Cormode, Cummings et~al.}]{kairouz2021advances}
Kairouz, P.; McMahan, H.~B.; Avent, B.; Bellet, A.; Bennis, M.; Bhagoji, A.~N.; Bonawitz, K.; Charles, Z.; Cormode, G.; Cummings, R.; et~al. 2021.
\newblock Advances and open problems in federated learning.
\newblock \emph{Foundations and Trends{\textregistered} in Machine Learning}, 14(1--2): 1--210.

\bibitem[{Krizhevsky, Nair, and Hinton(2009{\natexlab{a}})}]{kri2009cifar100}
Krizhevsky, A.; Nair, V.; and Hinton, G. 2009{\natexlab{a}}.
\newblock CIFAR-100 (Canadian Institute for Advanced Research).
\newblock \url{http://www.cs.toronto.edu/~kriz/cifar.html}.

\bibitem[{Krizhevsky, Nair, and Hinton(2009{\natexlab{b}})}]{krizhevsky2009cifar}
Krizhevsky, A.; Nair, V.; and Hinton, G. 2009{\natexlab{b}}.
\newblock Learning Multiple Layers of Features from Tiny Images.
\newblock \emph{CIFAR-10 (Canadian Institute for Advanced Research)}.

\bibitem[{Kurach et~al.(2019)Kurach, Lu{\v{c}}i{\'c}, Zhai, Michalski, and Gelly}]{kurach2019large}
Kurach, K.; Lu{\v{c}}i{\'c}, M.; Zhai, X.; Michalski, M.; and Gelly, S. 2019.
\newblock A large-scale study on regularization and normalization in GANs.
\newblock In \emph{International conference on machine learning}, 3581--3590. PMLR.

\bibitem[{Li et~al.(2017)Li, Yang, Song, and Hospedales}]{li2017deeper}
Li, D.; Yang, Y.; Song, Y.-Z.; and Hospedales, T.~M. 2017.
\newblock Deeper, broader and artier domain generalization.
\newblock In \emph{Proceedings of the IEEE international conference on computer vision}, 5542--5550.

\bibitem[{Li et~al.(2024)Li, Huang, Wang, and Shi}]{li2024global}
Li, H.; Huang, W.; Wang, J.; and Shi, Y. 2024.
\newblock Global and Local Prompts Cooperation via Optimal Transport for Federated Learning.
\newblock \emph{In Proceedings of the IEEE/CVF Conference on Computer Vision and Pattern Recognition}.

\bibitem[{Li et~al.(2020{\natexlab{a}})Li, Zhou, Xiong, and Hoi}]{li2020prototypical}
Li, J.; Zhou, P.; Xiong, C.; and Hoi, S.~C. 2020{\natexlab{a}}.
\newblock Prototypical contrastive learning of unsupervised representations.
\newblock \emph{arXiv preprint arXiv:2005.04966}.

\bibitem[{Li, He, and Song(2021)}]{li2020practical}
Li, Q.; He, B.; and Song, D. 2021.
\newblock Practical One-Shot Federated Learning for Cross-Silo Setting.
\newblock In \emph{Proceedings of the Thirtieth International Joint Conference on Artificial Intelligence (IJCAI-21)}.

\bibitem[{Li et~al.(2020{\natexlab{b}})Li, Sahu, Talwalkar, and Smith}]{li2020federated2}
Li, T.; Sahu, A.~K.; Talwalkar, A.; and Smith, V. 2020{\natexlab{b}}.
\newblock Federated learning: Challenges, methods, and future directions.
\newblock \emph{IEEE signal processing magazine}, 37(3): 50--60.

\bibitem[{Lin et~al.(2020)Lin, Kong, Stich, and Jaggi}]{lin2020ensemble}
Lin, T.; Kong, L.; Stich, S.~U.; and Jaggi, M. 2020.
\newblock Ensemble distillation for robust model fusion in federated learning.
\newblock \emph{Advances in Neural Information Processing Systems}, 33: 2351--2363.

\bibitem[{Liu et~al.(2023)Liu, Liu, Ye, Shen, Li, Jiang, and Li}]{liu2023fedlpa}
Liu, X.; Liu, L.; Ye, F.; Shen, Y.; Li, X.; Jiang, L.; and Li, J. 2023.
\newblock FedLPA: Personalized One-shot Federated Learning with Layer-Wise Posterior Aggregation.
\newblock \emph{arXiv preprint arXiv:2310.00339}.

\bibitem[{Luo et~al.(2021)Luo, Chen, Hu, Zhang, Liang, and Feng}]{luo2021no}
Luo, M.; Chen, F.; Hu, D.; Zhang, Y.; Liang, J.; and Feng, J. 2021.
\newblock No fear of heterogeneity: Classifier calibration for federated learning with non-iid data.
\newblock \emph{Advances in Neural Information Processing Systems}, 34: 5972--5984.

\bibitem[{Mai, Yan, and Pang(2024)}]{mai2024rflpa}
Mai, P.; Yan, R.; and Pang, Y. 2024.
\newblock Rflpa: A robust federated learning framework against poisoning attacks with secure aggregation.
\newblock \emph{arXiv preprint arXiv:2405.15182}.

\bibitem[{McMahan et~al.(2017)McMahan, Moore, Ramage, Hampson, and y~Arcas}]{McMahan17}
McMahan, H.~B.; Moore, E.; Ramage, D.; Hampson, S.; and y~Arcas, B.~A. 2017.
\newblock Communication-Efficient Learning of Deep Networks from Decentralized Data.
\newblock In \emph{AISTATS}.

\bibitem[{Mothukuri et~al.(2021)Mothukuri, Parizi, Pouriyeh, Huang, Dehghantanha, and Srivastava}]{mothukuri2021survey}
Mothukuri, V.; Parizi, R.~M.; Pouriyeh, S.; Huang, Y.; Dehghantanha, A.; and Srivastava, G. 2021.
\newblock A survey on security and privacy of federated learning.
\newblock \emph{Future Generation Computer Systems}, 115: 619--640.

\bibitem[{Neiswanger, Wang, and Xing(2013)}]{neiswanger2013asymptotically}
Neiswanger, W.; Wang, C.; and Xing, E. 2013.
\newblock Asymptotically exact, embarrassingly parallel MCMC.
\newblock \emph{arXiv preprint arXiv:1311.4780}.

\bibitem[{Netzer et~al.(2011)Netzer, Wang, Coates, Bissacco, Wu, and Ng}]{netzer2011reading}
Netzer, Y.; Wang, T.; Coates, A.; Bissacco, A.; Wu, B.; and Ng, A.~Y. 2011.
\newblock Reading digits in natural images with unsupervised feature learning.

\bibitem[{Nguyen, Munoz, and Jannesari(2024)}]{nguyen2024flora}
Nguyen, D.~P.; Munoz, J.~P.; and Jannesari, A. 2024.
\newblock Flora: Enhancing vision-language models with parameter-efficient federated learning.
\newblock \emph{arXiv preprint arXiv:2404.15182}.

\bibitem[{Nguyen et~al.(2022)Nguyen, Wang, Malik, Sanjabi, and Rabbat}]{nguyen2022begin}
Nguyen, J.; Wang, J.; Malik, K.; Sanjabi, M.; and Rabbat, M. 2022.
\newblock Where to begin? on the impact of pre-training and initialization in federated learning.
\newblock \emph{arXiv preprint arXiv:2206.15387}.

\bibitem[{Rao et~al.(2024)Rao, Zhang, Wu, Zhu, Sun, and Chen}]{rao2024privacy}
Rao, B.; Zhang, J.; Wu, D.; Zhu, C.; Sun, X.; and Chen, B. 2024.
\newblock Privacy inference attack and defense in centralized and federated learning: A comprehensive survey.
\newblock \emph{IEEE Transactions on Artificial Intelligence}.

\bibitem[{Sahu et~al.(2019)Sahu, Li, Sanjabi, Zaheer, Talwalkar, and Smith}]{sahu2019convergence}
Sahu, A.~K.; Li, T.; Sanjabi, M.; Zaheer, M.; Talwalkar, A.; and Smith, V. 2019.
\newblock Federated Optimization in Heterogeneous Networks.
\newblock In \emph{Proceedings of the Machine Learning and Systems (MLSys) Conference}.

\bibitem[{Tan et~al.(2022)Tan, Long, Liu, Zhou, Lu, Jiang, and Zhang}]{tan2022fedproto}
Tan, Y.; Long, G.; Liu, L.; Zhou, T.; Lu, Q.; Jiang, J.; and Zhang, C. 2022.
\newblock Fedproto: Federated prototype learning across heterogeneous clients.
\newblock In \emph{Proceedings of the AAAI Conference on Artificial Intelligence}, volume~36, 8432--8440.

\bibitem[{Tresp(2000)}]{tresp2000bayesian}
Tresp, V. 2000.
\newblock A Bayesian committee machine.
\newblock \emph{Neural computation}, 12(11): 2719--2741.

\bibitem[{Ulyanov, Vedaldi, and Lempitsky(2018)}]{inversion2}
Ulyanov, D.; Vedaldi, A.; and Lempitsky, V. 2018.
\newblock Deep image prior.
\newblock In \emph{Proceedings of the IEEE conference on computer vision and pattern recognition}, 9446--9454.

\bibitem[{Venkateswara et~al.(2017)Venkateswara, Eusebio, Chakraborty, and Panchanathan}]{venkateswara2017deep}
Venkateswara, H.; Eusebio, J.; Chakraborty, S.; and Panchanathan, S. 2017.
\newblock Deep hashing network for unsupervised domain adaptation.
\newblock In \emph{Proceedings of the IEEE conference on computer vision and pattern recognition}, 5018--5027.

\bibitem[{Wu et~al.(2024)Wu, Liu, Niu, Wang, Tang, and Zhu}]{wu2024fedlora}
Wu, X.; Liu, X.; Niu, J.; Wang, H.; Tang, S.; and Zhu, G. 2024.
\newblock FedLoRA: When Personalized Federated Learning Meets Low-Rank Adaptation.

\bibitem[{Xu, Tong, and Huang(2023)}]{xu2023personalized}
Xu, J.; Tong, X.; and Huang, S.-L. 2023.
\newblock Personalized federated learning with feature alignment and classifier collaboration.
\newblock \emph{arXiv preprint arXiv:2306.11867}.

\bibitem[{Yang et~al.(2023)Yang, Su, Li, and Xue}]{yang2023one}
Yang, M.; Su, S.; Li, B.; and Xue, X. 2023.
\newblock One-Shot Federated Learning with Classifier-Guided Diffusion Models.
\newblock \emph{arXiv preprint arXiv:2311.08870}.

\bibitem[{Yang et~al.(2024{\natexlab{a}})Yang, Su, Li, and Xue}]{yang2024exploring}
Yang, M.; Su, S.; Li, B.; and Xue, X. 2024{\natexlab{a}}.
\newblock Exploring One-Shot Semi-supervised Federated Learning with Pre-trained Diffusion Models.
\newblock In \emph{Proceedings of the AAAI Conference on Artificial Intelligence}, volume~38, 16325--16333.

\bibitem[{Yang et~al.(2024{\natexlab{b}})Yang, Su, Li, and Xue}]{yang2024feddeo}
Yang, M.; Su, S.; Li, B.; and Xue, X. 2024{\natexlab{b}}.
\newblock FedDEO: Description-Enhanced One-Shot Federated Learning with Diffusion Models.
\newblock \emph{arXiv preprint arXiv:2407.19953}.

\bibitem[{Yazdinejad et~al.(2024)Yazdinejad, Dehghantanha, Karimipour, Srivastava, and Parizi}]{yazdinejad2024robust}
Yazdinejad, A.; Dehghantanha, A.; Karimipour, H.; Srivastava, G.; and Parizi, R.~M. 2024.
\newblock A robust privacy-preserving federated learning model against model poisoning attacks.
\newblock \emph{IEEE Transactions on Information Forensics and Security}.

\bibitem[{Zhang et~al.(2022)Zhang, Chen, Li, Lyu, Wu, Ding, Shen, and Wu}]{zhang2022dense}
Zhang, J.; Chen, C.; Li, B.; Lyu, L.; Wu, S.; Ding, S.; Shen, C.; and Wu, C. 2022.
\newblock Dense: Data-free one-shot federated learning.
\newblock \emph{Advances in Neural Information Processing Systems}, 35: 21414--21428.

\bibitem[{Zhang et~al.(2020)Zhang, Sapra, Fidler, Yeung, and Alvarez}]{zhang2020personalized}
Zhang, M.; Sapra, K.; Fidler, S.; Yeung, S.; and Alvarez, J.~M. 2020.
\newblock Personalized federated learning with first order model optimization.
\newblock \emph{arXiv preprint arXiv:2012.08565}.

\bibitem[{Zhang et~al.(2023)Zhang, Xu, Yao, Zhang, Tian, and Wang}]{zhang2023federated}
Zhang, R.; Xu, Q.; Yao, J.; Zhang, Y.; Tian, Q.; and Wang, Y. 2023.
\newblock Federated domain generalization with generalization adjustment.
\newblock In \emph{Proceedings of the IEEE/CVF Conference on Computer Vision and Pattern Recognition}, 3954--3963.

\bibitem[{Zhou et~al.(2020)Zhou, Pu, Ma, Li, and Wu}]{zhou2020distilled}
Zhou, Y.; Pu, G.; Ma, X.; Li, X.; and Wu, D. 2020.
\newblock Distilled one-shot federated learning.
\newblock \emph{arXiv preprint arXiv:2009.07999}.

\end{thebibliography}

\newpage
\onecolumn
\section{Appendix}
\appendix
\subsection{Gaussian Naive Bayes classifier}
The close-formed solution of classifier Eq. \ref{classifier} could be obtained as follow:
% \small
\begin{align}
% \begin{flalign*}
\hspace{-4mm}
&p(y^{*}=j|f(x^{*}),\boldsymbol{\pi},\boldsymbol{\mu},\boldsymbol{\Sigma})=\frac{\pi_{j}\mathcal{N}(f(x^{*})|\mu^{j},\boldsymbol{\Sigma})}{\sum_{j^{\prime}}^{C}\pi_{j^{\prime}}\mathcal{N}(f(x^{*})|\mu^{j^{\prime}},\boldsymbol{\Sigma})} \notag \\
=&\frac{\pi_{j}\operatorname*{det}\left(2\pi\boldsymbol{\Sigma}\right)^{-\frac12}\operatorname{exp}\left(-\frac12\left(f(x^{*})-\mu^{j}\right)^{T}\boldsymbol{\Sigma}^{-1}\left(f(x^{*})-\mu^{j}\right)\right)}{\sum_{j^{\prime}}^C\pi_{j^{\prime}}\det\left(2\pi\boldsymbol{\Sigma}\right)^{-\frac12}\exp\left(-\frac12\left(f(x^*)-\mu^{j^{\prime}}\right)^T\boldsymbol{\Sigma}^{-1}\left(f(x^*)-\mu^{j^{\prime}}\right)\right)} \notag \\
=&\frac{\pi_j\exp\left(-\frac12f(x^*)^T\boldsymbol{\Sigma}^{-1}f(x^*)+(\mu^{j})^{T}\boldsymbol{\Sigma}^{-1}f(x^*)-\frac12(\mu^{j})^{T}\boldsymbol{\Sigma}^{-1}\mu^j\right)}{\sum_{j^{\prime}}^C\pi_{j^{\prime}}\exp\left(-\frac12f(x^*)^T\boldsymbol{\Sigma}^{-1}f(x^*)+(\mu^{j^{\prime}})^T\boldsymbol{\Sigma}^{-1}f(x^*)-\frac12(\mu^{j^{\prime}})^T\boldsymbol{\Sigma}^{-1}\mu^{j^{\prime}}\right)} \notag \\
=&\frac{\pi_j\exp\left((\mu^{j})^{T}\boldsymbol{\Sigma}^{-1}f(x^*)-\frac12(\mu^{j})^{T}\boldsymbol{\Sigma}^{-1}\mu^j\right)}{\sum_{j^{\prime}}^C\pi_{j^{\prime}}\exp\left((\mu^{j^{\prime}})^T\boldsymbol{\Sigma}^{-1}f(x^*)-\frac12(\mu^{j^{\prime}})^T\boldsymbol{\Sigma}^{-1}\mu^{j^{\prime}}\right)}.
% \end{flalign*}
\end{align}
Therefore, the weight $W\in \mathbb{R}^{C\times D}$ and the bias $b\in \mathbb{R}^{C}$ for the classifier can be expressed as:
\begin{align}
w_j=\boldsymbol{\Sigma}^{-1}\mu^j, b_j=\log\pi_j-\frac{1}{2}(\mu^j)^T\boldsymbol{\Sigma}\mu^j
\end{align}

\subsection{Additional Experimental Results.}
As shown in Table \ref{tab:pacs_officehome1}, we provide additional results on feature shit setting using DFKD methods DENSE \cite{zhang2022dense} and Co-Boosting \cite{dai2024enhancing}.
\vspace{-5pt}
\begin{table}[h]
\centering
\fontsize{9pt}{10pt}\selectfont % 将字体设置为 8pt，行间距为 10pt
\setlength{\tabcolsep}{4pt} % 将列间距设置为 12pt\
\begin{tabular}{c|c|cc|c|c|cc}
\toprule
        & Domain & DENSE & Co-Boosting &  & Domain & DENSE & Co-Boosting \\ \midrule
\multirow{5}{*}{PACS} 
        & P  & 40.29 & 49.18  & \multirow{5}{*}{OfficeHome}  & P  & 16.43 & 20.39  \\
        & A  & 22.62 & 33.47  &  & A  & 14.32 & 17.24  \\
        & C  & 28.42 & 32.64  &  & C  & 12.12 & 18.07  \\
        & S  & 38.77 & 40.52  &  & R  & 16.52 & 18.42  \\ \cline{2-4} \cline{6-8} 
        & Avg.  & 32.53 & 38.45  &  & Avg.  & 14.85 & 18.53  \\ \bottomrule
\end{tabular}
\caption{Performance on PACS and OfficeHome.}
\label{tab:pacs_officehome1}
\end{table}
\vspace{-10pt}

\subsection{Comparing with real feature statistics.}
We provide more results about deviation between global feature statistics captured by ours and the ground truth in Tabel \ref{tab:deviation1} and \ref{tab:deviation2}. Specifically, we conduct experiments on CIFAR100, SVHN using ResNet18 as the pre-trained model. Similar to the experiments performed on the CIFAR10 dataset, the deviations are not sensitive to the number of clients $M$ or the degree of heterogeneity $\alpha$. The accuracy evaluated on the test dataset remains stable across various scenarios.

\begin{table}[h]
\centering
% \small
% \scriptsize
\fontsize{9pt}{10pt}\selectfont % 将字体设置为 8pt，行间距为 10pt
\setlength{\tabcolsep}{4pt} % 将列间距设置为 12pt
\begin{tabular}{@{}c|ccc|ccc@{}}
\toprule
     &  \multicolumn{3}{c}{M=10}  & \multicolumn{3}{c}{M=50}  \\
     \midrule
     $\alpha$ & $\Delta\boldsymbol{\mu}$ & $\Delta\boldsymbol{\Sigma}$     & Acc & $\Delta\boldsymbol{\mu}$ & $\Delta\boldsymbol{\Sigma}$ & Acc     \\
     \midrule
0.05 & 2.03E-06 & 1.51E-05 & 39.95 & 1.63E-07 & 4.43E-06 & 39.95 \\
0.1  & 2.10E-06 & 1.53E-05 & 39.95 & 1.17E-07 & 3.08E-06 & 39.95 \\
0.5  & 1.53E-05 & 1.55E-05 & 39.95 & 9.25E-08 & 2.50E-06 & 39.95 \\
\bottomrule
\end{tabular}
\vspace{-5pt}
  \caption{The $L_2$ error between the FedCGS output and the real global feature statistics when using pre-trained ResNet18 on CIFAR100 (average 3 runs)} 
    \label{tab:deviation1}
    \vspace{-5pt}
\end{table}
\vspace{-10pt}

\begin{table}[h]
\centering
% \small
% \scriptsize
\fontsize{9pt}{10pt}\selectfont % 将字体设置为 8pt，行间距为 10pt
\setlength{\tabcolsep}{4pt} % 将列间距设置为 12pt\
\begin{tabular}{@{}c|ccc|ccc@{}}

\toprule
     &  \multicolumn{3}{c}{M=10}  & \multicolumn{3}{c}{M=50}  \\
     \midrule
     $\alpha$ & $\Delta\boldsymbol{\mu}$ & $\Delta\boldsymbol{\Sigma}$     & Acc & $\Delta\boldsymbol{\mu}$ & $\Delta\boldsymbol{\Sigma}$ & Acc     \\
     \midrule
0.05 & 3.67E-06 & 1.29E-05 & 57.77 & 1.22E-06 & 4.57E-06 & 57.77 \\
0.1  & 4.03E-06 & 1.45E-05 & 57.77  & 8.71E-07 & 3.64E-06 & 57.77 \\
0.5  & 3.07E-06 & 1.06E-05 & 57.77  & 1.36E-07 & 1.93E-06 & 57.77 \\
\bottomrule
\end{tabular}
\vspace{-5pt}
  \caption{The $L_2$ error between the FedCGS output and the real global feature statistics when using pre-trained ResNet18 on SVHN (average 3 runs)} 
    \label{tab:deviation2}
\vspace{-5pt}
\end{table}
\vspace{-10pt}

\end{document}